\documentclass[lnbip]{svmultln}

\newif\ifdraft\drafttrue
\draftfalse                                         %

\usepackage{hyperref}
\usepackage{relsize}
\usepackage{array}
\usepackage{paralist}
\usepackage{multirow}
\usepackage{booktabs}

\ifdraft
\usepackage{todonotes}
\else
\usepackage[disable]{todonotes}
\fi

\newcommand{\nop}[1]{}

\newcommand{\ourplatform}[0]{AI4EU Experiments Platform}

\begin{document}
\mainmatter              %
\title{Composing Complex and Hybrid AI Solutions}
\titlerunning{}  %
\author{%
Peter Sch\"uller\inst{1}
\and
Jo\~ao Paulo Costeira\inst{2}
\and
James Crowley\inst{3}
\and
Jasmin Grosinger\inst{4}
\and
F\'{e}lix Ingrand\inst{5}
\and
Uwe K\"ockemann\inst{4}
\and
Alessandro Saffiotti\inst{4}
\and
Martin Welss\inst{6}
}
\authorrunning{Peter Sch\"uller et. al}   %
\tocauthor{...}
\institute{
Knowledge-based Systems Group, Technische Universit\"at Wien, Vienna, Austria
\and
Institute for Systems and Robotics, T\'ecnico Lisboa, Lisbon, Portugal
\and
Univ.\ Grenoble Alpes, CNRS, Inria, Grenoble, France
\and
Center for Applied Autonomous Sensor Systems, \"Orebro University, Sweden
\and
LAAS-CNRS, University of Toulouse, France
\and
Fraunhofer Institute for Intelligent Analysis and Information Systems IAIS, Germany
}

\maketitle              %

\begin{abstract}
Progress in several areas of computer science has been enabled by comfortable and efficient means of experimentation, clear interfaces, and interchangable components,
for example using OpenCV for computer vision or ROS for robotics.
We describe an extension of the Acumos system towards enabling the above features for general AI applications.
Originally, Acumos was created for telecommunication purposes, mainly for creating linear pipelines of machine learning components.
Our extensions include support for more generic components with gRPC/Protobuf interfaces,
automatic orchestration of graphically assembled solutions including control loops, sub-component topologies, and event-based communication,
and provisions for assembling solutions which contain user interfaces and shared storage areas.
We provide examples of deployable solutions and their interfaces.
The framework is deployed at \url{http://aiexp.ai4europe.eu/} and its source code is managed as an open source Eclipse project.
\end{abstract}

\section{Introduction}

By the end of the $20^{\rm th}$ century the field of computer vision featured a vast repertoire of methods and algorithms, but suffered
from the lack of a common framework that would allow practitioners to
access these algorithms in a uniform way, and to compose them into
complex systems for their specific application.  In 2000, Intel released
the OpenCV library~\cite{opencv_library} as an infrastructure to make
optimized vision code easily available and reusable via standardized
interfaces.

A few years later the field of robotics was in a comparable situation:
hundreds of mature methods and algorithms were developed that could
potentially be used in different robotic hardware, but there was no easy
way to share these algorithms, to reuse them on different hardware, and
to compose them to build full robotic solutions.  In 2007 the Robot
Operating System (ROS) project was started~\cite{Quigley:2009tg}.
ROS provided a framework for robot software development where algorithms
could be wrapped in modular, reusable components, connected via
standardized interfaces.

Both OpenCV and ROS quickly became community standards, and they are
widely recognized for having produced a quantum leap in their respective
fields.  The ability to share, reuse and combine components allowed
researchers to easily build on previous results and to
compare competing techniques; it enabled companies to incorporate
existing solutions in their products; and it provided students with a
lowered entry barrier to experiment with advanced solutions.

Today, the field of Artificial Intelligence (AI) is in a similar
situation as computer vision and robotics were years ago.  We have a
large repertoire of mature methods and algorithms, but no standard
way to share them in a reusable format and no easy way to compose them
into complex solutions.  Some effective frameworks do exist that allow
the modularization and composition of machine learning components,
including Keras~\cite{keras}, PyTorch~\cite{pytorch},
Tensorflow~\cite{tensorflow} and Acumos~\cite{2018acumos}.  These,
however, are geared toward the use of data-driven, reactive
machine-learning components that are typically connected into simple,
linear pipelines.

What the field of AI needs today, in our opinion, is a more general framework
that can accommodate both data-driven and knowledge-based AI algorithms,
and that allows users to connect them in arbitrarily complex topologies.

In this paper, we propose such a framework that allows AI practitioners
to:

\begin{itemize}

\item embed their algorithms into a standard, portable format (docker
  containers);

\item interconnect these components using  standard interfaces
  (Protobuf and gRPC);

\item connect components in unrestricted topologies, including linear
  or branching pipelines, closed-loop systems or blackboard
  architectures;

\item accommodate both machine learning models and knowledge based
  components (such as logical reasoners, automated planners, constraint
  solvers, or ontological knowledge bases), allowing one to create
  hybrid AI solutions;

\item provide orchestration mechanisms to simplify the overall operation
  of a complex or hybrid solution.

\end{itemize}

The framework proposed in this paper is built on top of Acumos.  Acumos
is a state of the art system that already addresses the first two of the
above desired features: in this paper, we show how we have
extended Acumos beyond its initial scope in order to accommodate all the
remaining ones.  The resulting system is publicly available as the
AI4Experiments platform at \url{http://aiexp.ai4europe.eu/}.

The rest of this paper is organized as follows.  The next section
discusses some existing frameworks and their limitations in view of the
above desiderata.  Section~\ref{secRequirements} further elaborates
those desiderata.  Sections~\ref{secProposedApproach}
and~\ref{secPatterns} describe our proposed approach and the patterns
that it enables, respectively, while Section~\ref{secExamples} shows a
few case studies that illustrate those patterns.  Finally,
Section~\ref{secConclusions} concludes.

\section{Existing Frameworks and their Properties}
\label{secSOTA}

\begin{table}[tbp]\label{tblOverview}
\centering
\begin{tabular}{p{1.7cm}p{1.3cm}p{2cm}p{2cm}p{4cm}}
\textbf{Name} & \textbf{Focus} & \textbf{Designing Solutions} & \textbf{Running \& Deploying} & \textbf{Component Interface \& Communication} \\
\hline
\hline
\href{https://www.acumos.org/}{Acumos} & ML & GUI & k8s & Protobuf via REST, \mbox{no streaming} \\
\hline
\ourplatform & AI & GUI & k8s & Protobuf via gRPC, \mbox{streaming} \\
\hline
\href{https://www.ros.org/}{ROS} & Robotics & code & local & \href{http://wiki.ros.org/rosmsg}{rosmsg} via \href{https://www.omg.org/omg-dds-portal/}{DDS} \\
\hline
\href{https://opencv.org/}{OpenCV} & CV & code & library & C++ library \\
\hline
\href{https://www.kubeflow.org/}{Kubeflow} & ML & Python DSL/GUI & k8s & Cloud Storage, no~interface \\
\hline
\href{https://www.h2o.ai/}{H2O.ai} & Parallel ML & GUI or code & various & distributed key/value~store \\
\hline
DL \mbox{frameworks} & DL & code & library & Python \\
\hline
commercial platforms & ML & code & k8s & Protobuf{\,+\,}gRPC\,(Google), Smithy (Amazon) \\
\hline
\hline
\end{tabular}
\caption{%
Overview of frameworks and platforms for modular assembly of AI applications.
Abbreviations:
ML = Machine Learning, AI = Artificial Intelligence,
CV = Computer Vision, DL = Deep Learning,
k8s = Kubernetes, %
GUI = Graphical User Interface,
DL frameworks such as \href{https://www.tensorflow.org/}{TensorFlow} and \href{https://keras.io/}{Keras},
commercial platforms such as Google Vertex and Amazon Sagemaker.%
}
\end{table}

We next give an overview of frameworks for modular assembly of AI applications.
We consider frameworks that provide an interface for component-component communication on a higher level than simply providing network communication.
Table~\ref{tblOverview} provides an overview of frameworks that are discussed in detail in the following sections.
Finally we give a coarse overview of further frameworks.

\subsection{Acumos}

Acumos \cite{2018acumos} is a software framework created by AT\&T
for the needs of big telecommunication providers.
It was initially conceived purely for linear machine learning pipelines,
i.e., sequences of components with an acyclic information flow
from one or more sources to one or more sinks.

Acumos contains a graphical web interface for assembling \emph{Solutions} from \emph{Components}
and a marketplace where components and solutions can be shared with other users or made publicly available.
A Component is a software artifact that has an interface in terms of Google Protocol Buffer (Protobuf) \cite{protobuf} definition.
Protobuf definitions permit to define message data types and services.
A Protobuf service contains RPC calls with exactly one input and one output message data type.
Acumos creates a \emph{Port} for each input and for each output of an RPC call,
and ensures that only Ports with matching types are connected in a solution.

An important aspect of Acumos solutions is that each component is a passive server
and that the solution becomes executed by means of an \emph{orchestrator} component which passes messages between components in the correct order.
Acumos is a modular system and contains many APIs with possibilities
for plugging in, e.g., new component types or new orchestrators.

\subsection{ROS}
ROS~\cite{Quigley:2009tg} is a very popular software development framework in the robotic community. Programs written in ROS are
\emph{nodes} which communicate through an asynchronous publish/subscribe mechanism over \emph{topics}. One node can advertise a \emph{topic}
(e.g. the position of the robot) to contain some data, whose format is defined in a \emph{msg} file (e.g. three floats: x, y, theta) , and
then publish this data at will. Other nodes can subscribe to this \emph{topic} and specify a callback which will be called when a new topic
value (e.g. a new position of the robot) is published. Another mechanism, \emph{service}, is also provided to make a synchronous remote process call
(RPC) to a \emph{server} advertising the services (with the type of data passed as arguments and returned by the call defined in a \emph{srv}
file). This can be used for example to have a server node to provide a locate service which given the x, y (e.g. 10.3, 4.0) position of the robot, returns
the name of the room in which it is (e.g. "kitchen").

These two mechanisms and the definition in \emph{msg} of commonly used data structure in robotics (e.g. odometry, images, point cloud, etc)  has led to a very dynamic ecosystem of
nodes using topics produced by others and providing topics to others. As a result, most robotic equipment manufacturers provide ROS nodes to
control their robots, sensors, effectors, and most main stream robotic algorithms (SLAM, navigation, etc) have a number of implementation
available in ROS. Just sharing some common data structure definition and providing a versatile and simple data communication mechanism led
the robotics community to share data, results and algorithms like never before and enable newcomers to get involved and active in complex
robotic experiments with little initial programming investment.

ROS is now in its second installment  which addresses some of the shortcomings of the first version: DDS is now used as middleware; no more ROS core
centralizing the book keeping of publishers and subscribers of topics,  servers and clients of services;  multi/mono CPU deployment; etc). ROS is a clear and successful example of what can be achieved with just sharing data structure definition and a simple communication mechanism.

\subsection{OpenCV}
OpenCV \cite{opencv_library} is an open source computer vision and machine learning software library  for computer vision applications that is cross-platform and free for use under the open-source Apache 2 License, allowing easy use for  commercial applications. OpenCV was originally developed in the late 1990's by Gary Bradski as an Intel Research initiative to advance CPU-intensive applications. The first alpha version of OpenCV was released to the public at the IEEE Conference on Computer Vision and Pattern Recognition in 2000. Development and support was taken over by Willow Garage in the early 2000s, and Version 1.0 was released in 2006. A second major release in October 2009 included major changes to the C++ interface, and other improvements, with support GPU acceleration added in 2011.
In August 2012, support for OpenCV was taken over by a non-profit foundation OpenCV.org, which currently maintains a developer and user web site. Development is provided by an independent Russian team supported by commercial corporations, with Official releases approximately every six months.
The most recent version has more than 2500 optimized algorithms, including both classic and state-of-the-art computer vision and machine learning algorithms with more than 47,000 active users and estimated downloads exceeding 18 million. The library is used extensively by companies, universities, research groups and governmental bodies.

OpenCV has C++, Python, Java and MATLAB interfaces and supports Windows, Linux, Android and Mac OS. It is optimized for real-time vision applications and takes advantage of MMX and SSE instructions when available. Full-featured CUDA and OpenCL interfaces are under development. There are over 500 algorithms with about 10 times as many functions that compose or support the algorithms. OpenCV is written  in C++ and has a template interfaces that work seamlessly with STL containers.

Public availability of OpenCV and its rich collection of functionalities available in a uniform programming framework available for several platforms has been an important factor in the rapid growth of commercial and industrial use of Computer vision over the last decade.
\subsection{Kubeflow}

KubeFlow is an open source machine learning platform originally created by Google to simplify the management of deep learning workflows by leveraging the features of Kubernetes. The workflows can be designed using a Python based DSL or a Web-GUI. The nodes of a workflow are Kuberentes pods that communicate only by input and output files. The nodes have no services defined and do not communicate directly. The files are exchanged via cloud storage that is provided outside the workflow definition. The workflow basically defines the dependencies on other nodes (or tasks), very much like in a makefile. It is then up to the orchestrator (workflow engine) to find the best order for execution and level of parallelism. If all preconditions for a node are met, the pod is started and the task ends when the pods has written its output files and terminates. Then the pods for the tasks depending on it are started.

Here is a small example: task A is data cleaning, task B is model training and depends on task A.

The workflow engine reads the dependencies and concludes that task A must be run before task B.

The pod of task A is started, it reads the data files from cloud storage, cleans the data and writes new files with cleaned data to the cloud storage and terminates.

Then the pod of task B is started, it reads the files with cleaned data form cloud storage and stores the trained model somewhere.

\subsection{H2O.ai}

H2O.ai%
\footnote{\url{https://www.h2o.ai/}}
is an open source framework for ML with a focus on parallelization and scaling up ML in practice.
It can be deployed in Map/Reduce cloud infrastructures of all popular providers, on Hadoop, Spark, and locally.
Several popular ML and Data Science algorithms are provided as Map/Reduce implementations.
Custom algorithms can be implemented as well in Python, R, Scala, or Java.
These languages or the H2O Flow GUI is used to design H2O ML applications.
H2O has AutoML capabilities to discover the best algorithm for a given task.

\subsection{Deep Learning Frameworks}

Tensorflow~\cite{tensorflow} is a library for
developing ML applications and algorithms, supported by
dedicated hardware, if present (e.g., GPUs).
It offers a low-level and a high-level API in several
languages (e.g., Python) and is not specific
to neural network applications.
Keras~\cite{keras} and pytorch~\cite{pytorch}
are libaries for more high-level development of ML applications,
where Keras (based on Tensorflow) is focused on neural networks.
Common to these and other frameworks is,
that they offer a Python API to assemble by means of
writing a Python program a ML application
in a comfortable way. Below that Python API are efficient low-level implementations of ML algorithms
that can operate on GPUs and on large-scale compute clusters.
Communication between the algorithm parts is managed by the library.

\subsection{Commercial Platforms}

Google Vertex AI~\footnote{\url{https://cloud.google.com/vertex-ai}} describes itself as being a unified AI platform that facilitates building, deploying and scaling of ML models. That means that it brings together Google cloud services for building ML under one UI and API. It integrates ML frameworks such as TensorFlow, PyTorch and scikit-learn as well as frameworks via custom containers. Vertex can do data preparation (ingest, analyze, transform) and then be used to train, model, evaluate, deploy, and predict. Google Vertex AI is cloud-based, so to work with it, one logs in to the Google Cloud Platform, where a new project can be created. In the Google Cloud Shell (or locally, if preferred)  a storage bucket is created for storing saved model assets for a training job. Next Docker files and containers are to be created. Training code can be written in Python, for example, using TensorFlow, but other open source frameworks or custom frameworks are possible, as mentioned above. The Docker container can now be built and tested locally and finally pushed to the Google Container Registry. There are two options for training models in Vertex: AutoML or Custom training. In the Google Cloud web-interface one can create the training job together with entering the parameters and the deployed model, as well as selecting the Docker container built in the previous step. Finally an endpoint of the trained model can be created which can be used to get predictions on the model.

Amazon Sagemaker~\footnote{\url{https://aws.amazon.com/sagemaker/}} is infrastructure, tools and managed workflows for building, training and deploying ML models. Business analysts can use the visual interface \emph{Sagemaker Canvas} and can prepare data, train models and create predictions without having to write code. For data scientists, Amazon Sagemaker offers an IDE for the ML life cycle. The so-called \emph{Studio Notebooks} can access data from both structured and unstructured data sources which is then prepared. Next ML models are built. Built-in ML algorithms can be used or own algorithms. Frameworks such as \emph{TensorFlow} and \emph{PyTorch} are supported. Then the ML model is trained. When deploying the ML model, it can be continuously monitored --- model and concept drifts can be detected and alerted, and key metrics can be collected and viewed. MLOps Engineers can streamline the ML lifecycle. They can build CI/CD pipelines to reduce management overhead, automate ML workflows, that is, accelerate data preparation and model building, training and experiments. Amazon Sagemaker Pipelines are a feature to help automate and orchestrate different steps of the ML workflow such as data loading and transformation, model building,  training and tuning. Such pipelines support processing a large amount of training data, run large-scale experiments, build and re-train models at various scales. Workflows can be re-used and shared.

\subsection{Coarse Overview of Other Frameworks}

\subsubsection{Containerization and Virtualization}

Several of the frameworks described above are based on generic virtualization and containerization technology
such as Docker, VMware, and \href{https://kubernetes.io}{Kubernetes}.
This technology allows for creating images of operating systems with prepackaged software.
These images are ready to run on computers with the respective host software
without the need for specific setup operations and sometimes even without the need to run on the correct
hardware architecture.
Furthermore containerization and virtualisation permits easy restarting from a known state of an image
and comfortable switching between versions of images in a running deployed application.

While these technologies are often an important part of the infrastructure for modular AI applications,
we do not consider them separately in the following discussion
because containerization and virtualization does not provide two essential ingredients of modular AI applications:
(a) a high-level interface language for describing communication formats between components, and
(b) a possibility to compose components into applications without changing the components.
These two ingredients are provided by containerization and virtualization technology only on the low level network layer.

\subsubsection{Machine Learning}

Scikit-learn~\footnote{\url{https://scikit-learn.org}} is an open source machine learning library for Python. It supports both supervised and unsupervised learning and provides tools for model fitting, data preprocessing, model selection and evaluation. It provides built-in ML algorithms and models, called estimators. It is possible to chain pre-processor and estimators in a pipeline. This term, pipeline, is understood as a sequential application of transforms and a final estimator.

Weka~\footnote{\url{https://www.cs.waikato.ac.nz/ml/weka/}} too is open source and ML but is a collection of ML algorithms in Java that can be used for classification, regression, clustering, visualization and more. It supports DL too.

\subsubsection{Specific Kubernetes-based Frameworks}
AWS Proton~\footnote{\url{https://aws.amazon.com/proton/}} is a tool from Amazon for automating the management of containers and do serverless deployments based on OpenAPI interfaces.
Lightbend Akka Serverless~\footnote{\url{https://www.lightbend.com/akka-serverless}} %
is based on first creating data artifacts using a Protobuf API and then writing code which operates on these artifacts.
Durable storage of these artifacts is handled automatically with the goal of low latency ``real-time" performance and without the need to have any knowledge about databases.

Both AWS Proton and Lightbend Akka Serverless require components to know in advance which other services they will access.
Therefore, composing arbitrary solutions from existing components without modification of the components is difficult.

\subsubsection{Natural Language Processing (NLP)}

Several popular NLP frameworks exist.
The Natural Language Toolkit, nltk~\footnote{\url{https://www.nltk.org/}},
is an open source platform for building NLP programs with Python.
Nltk provides interfaces to over 50 corpora and lexical resources as well as text processing libraries for classification, tagging, parsing, semantic reasoning and more.
The General Architecture for Text Engineering, Gate~\footnote{\url{https://gate.ac.uk/}},
is an open software toolkit for solving text processing problems.
It contains a graphical user interface
and an integrated development environment for language processing components.
Apache UIMA~\footnote{\url{https://uima.apache.org/}} (Unstructured Information Management Applications) is a software system for analyzing large volumes of text to discover knowledge that can be of relevance. UIMA wraps components in network services and includes scalability provisions
by replicating modular processing pipelines over a cluster of networked nodes.

\subsubsection{Catalogs and Package Managers}
Further important frameworks are related to this work
but excluded from the overview because they have a different focus.
\href{https://openml.org/}{OpenML} is a catalog/documentation platform
for ML datasets, algorithms, and evaluation results.
\href{https://www.anaconda.com}{Anaconda} is a repository of (pre-built) AI software packages
with a focus on enabling replicable installations of lists of packages in mutually compatible versions.

\newcommand{\tabFeatures}[0]{
\begin{table}[tbp]
\smaller
\centering
 \begin{tabular}{p{3.8cm} p{3.8cm} p{3.8cm}}
 \toprule
 \textbf{Feature} &
    \textbf{Requirement} &
        \textbf{Advantage or Pattern} \\ \midrule
 Container Specification &
    - &
        \multirow{2}{3.8cm}{lower entrance barrier, broad reusability of components} \\[2pt] \cline{1-1}
 Easy Deployment \raisebox{2pt}{\strut} &
    - &  \\ \midrule
 Model Initializer Component &
    - & more generic components \\ \midrule
 Shared Filesystem Component &
    - & bringing data to components more efficiently \\ \midrule
 Generic Orchestrator &
    cyclic topologies &
        control loops, user interfaces \\ \midrule
 Streaming RPC &
    non-batch dataflow &
        user interfaces, sub-components \\
 \bottomrule
 \end{tabular}
 \caption{Novel features of the \ourplatform\ and their interaction with requirements and resulting advantages/enabled patterns.}\label{tab:features}
\end{table}
}

\section{Requirements for a Modular Hybrid AI Framework}
\label{secRequirements}

There is a need for a broad component-based reusable approach for Artificial Intelligence.

We consider as AI everything that
\begin{itemize}
    \item
        has a "model" of reality (learned, manually written, or combinations), and
    \item
        performs "reasoning" on that model (computation such as prediction, inference, learning, verification, search).
\end{itemize}

\subsection{Requirements on types of models}

Models can be static or they can be updated during reasoning. They might take into account uncertainty and probability.
The framework shall provide the possibility for using pretrained models as well as training and predicting with models within one application.

Moreover, models can be modularly constructed from other models.
This possibility is not limited to typical ensemble predictors but also applies to,
e.g., methods for explaining or verifying the predictions of other (black-box or white-box) models.

Reasoning with models can be deterministic or randomized, online or offline, batched or single-shot.

\subsection{Requirements on communication between models}
Models are not used in isolation, they can be connected to other models.
Moreover, models can interact with components that connect with the real world (with humans or with other AI agents).

Connections among models can lead to multiple cycles across components.
This is especially common in robotic applications where multiple hierarchically nested control cycles are frequently used.

Another aspect on communication between models is the data volume:
communication can be low-volume (e.g., location information for a robot)
or high-volume (e.g., a whole dataset of images with labels for learning).

\section{The \ourplatform}
\label{secProposedApproach}

\tabFeatures
To address the needs we described previously,
we propose the \ourplatform\ which extends the Acumos system in several ways.
We chose Acumos as a basis because it is open source under a permissive license,
uses a modular microservice architecture,
provides a catalog and private sharing of components,
and because it provides a graphical editor for building solutions.

We next describe how we propose to transform Acumos into a platform for Hybrid AI applications in general.
Table~\ref{tab:features} gives a structured overview.
\subsection{Container Specification}
    We define a simplified {\bfseries format for components}: all components are Docker containers that must have a gRPC server listening on port 8061 and can have a webserver listening on port 8062.
    The webserver can provide information about the component or it can be the main aspect of a component, i.e., if the component is a graphical user interface (GUI).

    The rationale for that is to make authoring of components easier,
    to enable uniform deployment of all components,
    and to have cleary defined interfaces for all components.

    This improves component re-use and interoperability between components of different authors.

\subsection{Easy Deployment}
    We {\bfseries simplify deployment} of solutions:
    we provide a button for downloading a ZIP file which contains
    (i) a script for deploying all components and an orchestrator component in a kubernetes environment, and
    (ii) a script for interacting with the orchestrator component for executing the solution.

    The deployment script requires as input just the namespace where the whole solution shall be deployed.
    The orchestrator client script can start and stop orchestration and it makes orchestration event logs accessible.

    The goal of this extension is to make obtaining, deploying, and running a solution as easy as possible.
    Event logs are helpful for seeing how a solution is orchestrated and diagnosing potential problems.

\subsection{Model Initializer Component}
    We provide a new {\bfseries component type for initializing other components},
    e.g., with machine learning models or knowledge bases.

    This component is not a deployed Docker container but it represents a changes of the deployment of all components that are connected to the Model Initializer component.

    This makes component initialization explicit.
    It also facilitates more generic components,
    because the (learned or manually curated) AI model inside a component does not need to be fixed---it can be initialized by an initializer component.

\subsection{Shared Filesystem Component}
    We provide a new {\bfseries component type which represents shared file systems}.
    Each component can obtain access to such a shared file system
    by means of an explicit link in the solution.

    Like the Model Initializer, this component is not a deployed Docker container but an explicit representation of a change to the solution deployment.

    Shared filesystems permit data-intense applications to access the same data
    without passing it over gRPC messages.
    Moreover, it permits to \emph{execute a solution where the data is}
    by providing existing shared volumes in kubernetes
    to components of a solution for processing.

\subsection{Generic Orchestrator}
    We relax many constraints on the allowed topologies of solutions
    by means of a {\bfseries new orchestrator} software
    that is able to run applications with {\bfseries topologies that contain cycles}.

    The orchestrator is very general and based on using multiple threads instead of computing an execution order.
    Therefore, it can deal with any topology as long as connections between components respect interfaces.

    An important rationale for the new orchestrator is the need for feedback cycles
    in many AI applications, in particular control loops, e.g., in robotics applications.

\subsection{Streaming RPC}
    We permit {\bfseries streaming RPC} both for input and output of RPC calls.
    Streaming RPC starts a call and then permits to stream in (or out) a variable number of messages.
    Streaming is the gRPC word for event-based interaction.
    Importantly, there can be arbitrary delays between messages.
    An RPC immediately receives each input message as soon as it is sent by the previous component,
    and an RPC can decide when to send output messages on a stream, and how many messages before the RPC is closed.

    This enables asynchronous communication,
    components using other components as sub-components, and cyclic information flow.
    In particular, user interfaces can trigger computations based on user actions (events)
    and display results from computations of other components.

\section{Enabled Patterns}
\label{secPatterns}

These extensions enable the following patterns for composing applications.
These patterns are not possible using the original Acumos software.

\subsection{Graphical User Interfaces that interact with components}
\label{secPatternGUI}

This pattern permits a component to act as a graphical user interface (GUI)
which sends events to a solution and displays the results of that solution.
Events are emitted via streaming output, results are ingested via streaming input.
For each type of result to be visualized, the GUI component can have a separate
RPC.
Multiple types of results can be visualized at different rates this way.
The solution and the components of the solution that receive events from the GUI
and send results to the GUI do not need to be aware that they will be connected to a GUI.

An implemented example of a GUI is the Sudoku Design Assistant GUI%
\footnote{\url{https://github.com/peschue/ai4eu-sudoku/tree/streaming/gui}}
which has the following interface.
\begin{verbatim}
service SudokuGUI {
  rpc requestSudokuEvaluation(Empty)
      returns(stream SudokuDesignEvaluationJob);
  rpc processEvaluationResult(stream SudokuDesignEvaluationResult)
      returns(Empty);
}
\end{verbatim}
The first RPC emits a job for each user event that requires a new evaluation.
The second RPC displays results.
The Sudoku topology is described in Section~\ref{secSudokuExample}.
Another example that uses the GUI pattern is the maze planner, described in Section~\ref{secMazePlannerExample}.

\subsection{Sub-components}
\label{secPatternSubcomponent}

\begin{figure}[tbp]
    \centering
    \includegraphics[scale=0.15]{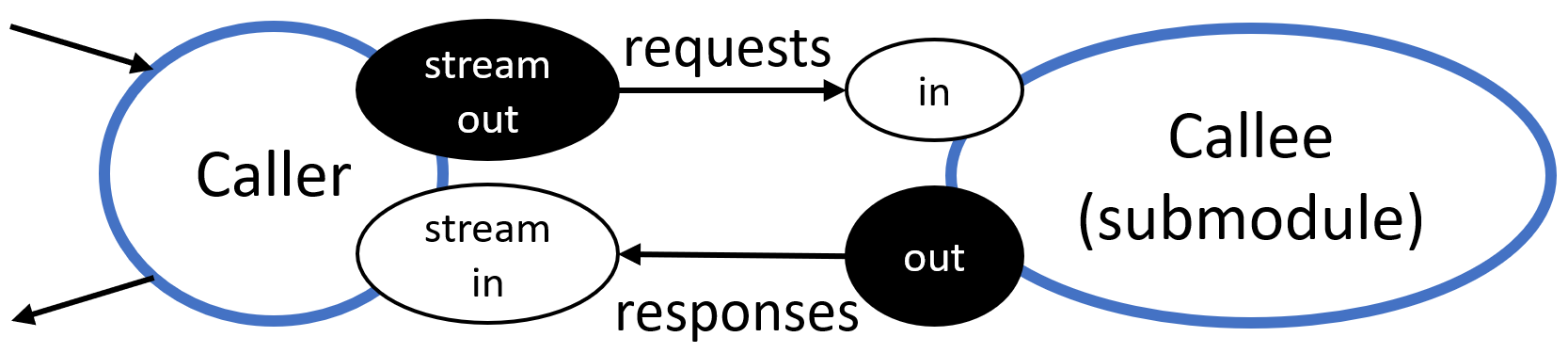}
    \caption{Subcomponent pattern. The arrows on the left indicate an arbitrary interaction of the caller with other components.}
    \label{figSubcomponent}
\end{figure}
This pattern permits a component to use the functionality of another component for computing a result,
illustrated in Figure~\ref{figSubcomponent}.
Calling a sub-component is achieved by emitting requests on a stream output RPC, ingesting results using a stream input RPC, and connecting caller and callee in a cyclic topology.
The caller can call the callee once or multiple times.
The sub-component does not need to provide a specific interface to be `callable' in that way.
The caller may call one component and ingest results from another component,
so the `sub-component' may actually be a `sub-solution`.

\begin{example}

An implemented example of a subcomponent is the Answer Set Solver of the Sudoku Solution%
\footnote{\url{https://tinyurl.com/368c3t6w}}%
$^,$%
\footnote{\url{https://github.com/peschue/ai4eu-sudoku/tree/streaming}}
which is a generic component with the gRPC interface
\begin{verbatim}
service OneShotAnswerSetSolver {
    rpc solve(SolverJob) returns (SolveResultAnswersets);
}
\end{verbatim}
where the input \verb|SolverJob| indicates how many answers are of interest
and \verb|SolveResultAnswersets| contains all answers.
The Sudoku Design Evaluator,%
\footnote{\url{https://github.com/peschue/ai4eu-sudoku/tree/streaming/evaluator}}
which is using the ASP Solver as a subcomponent,
has the gRPC interface
\begin{verbatim}
service SudokuDesignEvaluator {
    rpc evaluateSudokuDesign(SudokuDesignEvaluationJob)
        returns (SudokuDesignEvaluationResult);
    rpc callAnswersetSolver(Empty) returns(stream SolverJob);
    rpc receiveAnswersetSolverResult(stream SolveResultAnswersets)
        returns(Empty);
}
\end{verbatim}
where \verb|evaluateSudokuDesign| is the way the GUI uses the Design Evaluator,
the RPC \verb|callAnswersetSolver| emits requests to the ASP Solver,
and the RPC \verb|receiveAnswersetSolverResult| ingests the results.

If a large number of answers is of interest,
the solver can stream out solutions using the following interface.
\begin{verbatim}
service OneShotAnswerSetSolver {
    rpc solve(SolverJob) returns (stream SolveResultAnswerset);
}
\end{verbatim}
Here, each output in the stream contains a single result.
\hfill\strut\qed
\end{example}

\subsection{Control Loops}
\label{secPatternControl}

Topologies can contain cycles:
the output of a component is passed to a component that directly
or indirectly feeds back a result into the same component.
Different from the previous two patterns,
in this pattern there is no notion of a singular computation `event' in the solution.
Instead, the cycle periodically passes messages around in order to
realize a control loop,
where a controller component receives sensor input from the environment
and emits output to influence the environment.
Importantly, sensor input can be transferred at a different rate than controller output,
if desired.
Additionally, the goal of the controller can be updated asynchronously using another stream.
Moreover, multiple cycles can exist, e.g., a slow high-level controller
that uses reasoning to set low-level goals, which are fed into a fast low-level controller that receives sensor information and sends actuator signals to a robot in the environment.

A control loop topology is used by the Maze planner example, see Section~\ref{secMazePlannerExample}: multiple cycles exist: the executor performs actions in the simulator and needs to trigger re-planning if an action fails, leading to further actions and potentially re-plannings.

\section{Example Applications / Case Studies}
\label{secExamples}

\subsection{Sudoku Tutorial}
\label{secSudokuExample}

{%
\newcommand{\myevaluationjob}{\includegraphics[scale=0.3]{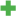}}%
\newcommand{\myevaluationresult}{\includegraphics[scale=0.3]{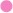}}%
\newcommand{\myempty}{\includegraphics[scale=0.3]{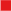}}%
\newcommand{\mysolverjob}{\includegraphics[scale=0.3]{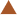}}%
\newcommand{\mysolverresult}{\includegraphics[scale=0.3]{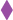}}%
\newcommand{\figureSudoku}[0]{
\begin{figure}[tbph]
    \centerline{%
    \begin{tabular}{@{}l@{ }l@{}}
    \includegraphics[scale=0.35]{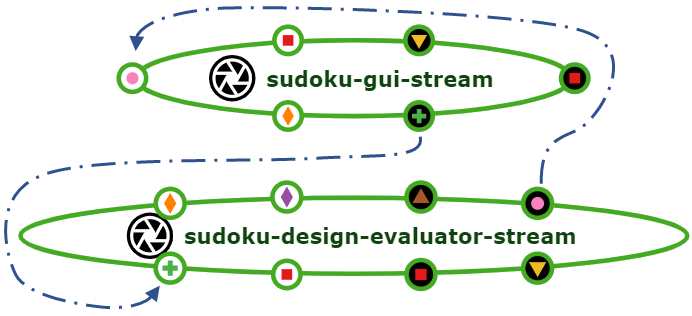}
    &
    \includegraphics[scale=0.35]{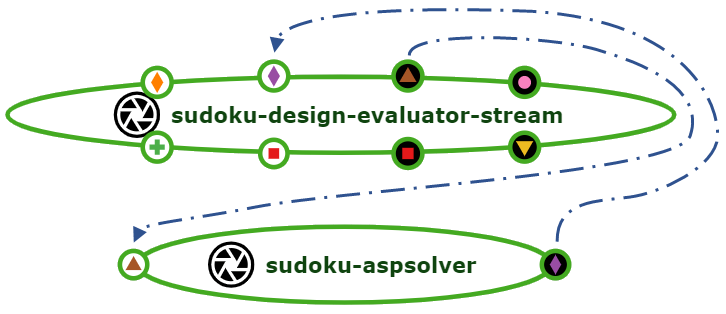}
    \end{tabular}%
    }
    {%
    \scriptsize%
    \begin{tabular}{@{}l@{}}
    \small{}GUI Interface: \\
    \tt{} rpc requestSudokuEvaluation(\myempty Empty) returning (stream \myevaluationjob SudokuEvaluationJob);\\
    \tt{} rpc processEvaluationResult(stream \myevaluationresult SudokuEvaluationResult) returning \myempty Empty); \\[4pt]
    \small{}Design Evaluator Interface: \\
    \tt{} rpc evaluateSudokuDesign(\myevaluationjob SudokuEvaluationJob) returning (\myevaluationresult SudokuEvaluationResult); \\
    \tt{} rpc callAnswersetSolver(\myempty Empty) returning (stream \mysolverjob SolverJob); \\
    \tt{} rpc receiveAnswersetSolverResult(stream \mysolverresult SolveResultAnswersets) returning (\myempty Empty); \\[4pt]
    \small{}ASP Solver Interface: \\
    \tt{} rpc solve(\mysolverjob SolverJob) returning (\mysolverresult SolveResultAnswerset);
    \end{tabular}%
    }
    \caption{Sudoku Tutorial connections (above) and Protobuf interfaces (below). Ports with white (resp., black) background are input (resp., output) ports.
    The {\tt sudoku-design-evaluator-stream} component is a single component which is
    depicted twice for presentation reasons.%
    }
    \label{figSudokuTutorial}
\end{figure}%
}
\figureSudoku

The Sudoku Tutorial is a solution comprising fully open-sourced components%
\footnote{\url{https://github.com/peschue/ai4eu-sudoku/}}
with the purpose of helping others to create assets and solutions.
It consists of a web interface (GUI) where one can configure a partial Sudoku grid,
and with each change the Design Evaluator component computes up to two solutions
to the Sudoku and returns the common digits in the grid to the GUI. If there is no
solution, a minimal repair for the fixed digits is computed and returned to the GUI.
The Design Evaluator performs these computations using a generic third component,
the Clingo~\cite{gebser2011potassco} Answer Set Solver.
Figure~\ref{figSudokuTutorial} shows the components as they are displayed in the
graphical user interface of the \ourplatform, including their connections and Protobuf interfaces.
This Tutorial contains streaming for the purpose of sending user events to the Design Evaluator
and for sending display updates to the user interface,
moreover streaming is used for integrating the Answer Set Solver component
as a subcomponent to the Design Evaluator.
Hence the tutorial uses the GUI and Sub-component patterns, see Sections~\ref{secPatternGUI} and~\ref{secPatternSubcomponent}, respectively.

For a quick start into developing suitable components,
the repository contains a script {\tt helper.py} which provides the following functionalities:
\begin{inparaenum}[(i)]
\item running each of the three components outside of docker;
\item orchestrating locally running components with a hardcoded (very short) orchestrator script;
\item building docker images for each of the three components;
\item running, stopping, and following these docker images in a local docker installation; and
\item pushing docker images to a docker repository.
\end{inparaenum}

The complete Sudoku example%
\footnote{\url{https://tinyurl.com/26wvv4j4}}
can be deployed using the ``Deploy to Local'' functionality
and executing {\tt kubernetes-client-script.py -n NAMESPACE} in a kubernetes environment where
{\tt NAMESPACE} is an empty namespace for deployment.
This script waits for all containers to run in the kubernetes namespace
and then starts the orchestration and displays orchestration events.

For more details see the detailed walk-through Sudoku tutorial on YouTube.%
\footnote{\url{https://youtu.be/gM-HRMNOi4w}}

}%

\subsection{Planning framework and control circuit}
\label{secMazePlannerExample}

\begin{figure}[tbph]
    \centering
    \includegraphics[scale=0.32]{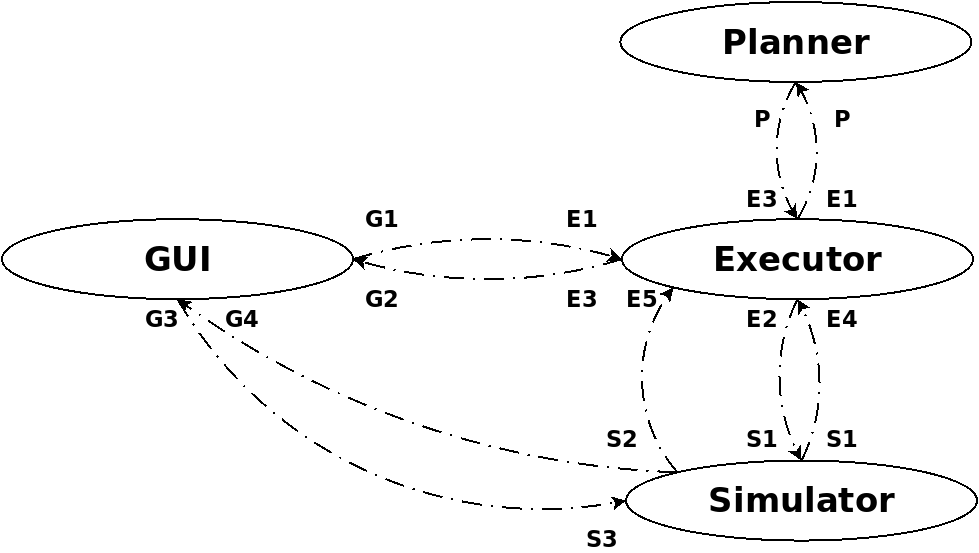}
    {%
    \scriptsize%

    \begin{tabular}{c|@{}l@{}}
       & \small{} GUI Interface: \\
       \hline
    G1 & \tt{} rpc requestTask(Empty) returns(Goal); \\
    G2 & \tt{} rpc processTaskResult(Result) returns(Empty); \\
    G3 & \tt{} rpc getState(Empty) returns(State); \\
    G4 & \tt{} rpc visualizeState(State) returns(Empty); \\[4pt]
       & \small{} Simulator Interface: \\
       \hline
    S1 & \tt{} rpc doAction(Action) returns (Result); \\
    S2 & \tt{} rpc getState(Empty) returns (State); \\
    S3 & \tt{} rpc setState(State) returns (Empty); \\[4pt]
       & \small{} Planner Interface: \\
       \hline
    P  & \tt{} rpc plan(Problem) returns (Solution);\\[4pt]
       & \small{} Executor Interface: \\
       \hline
    E1 & \tt{} rpc assembleProblem(Goal) returns (Problem); \\
    E2 & \tt{} rpc doNextAction(Empty) returns (Action); \\
    E3 & \tt{} rpc processPlanningResult(Solution) returns (Result); \\
    E4 & \tt{} rpc processActionResult(Result) returns (Empty); \\
    E5 & \tt{} rpc processState(State) returns (Empty); \\[4pt]
    \end{tabular}%
    }
    \caption{Maze planner connections (above) and Protobuf interfaces (below). Connection arrows are annotated with references to the interfaces below. Symbols at the origin/destination of an arrow indicate the output/input of the corresponding RPC is used.}
    \label{figMazePlanner}
\end{figure}

The maze-planner example\footnote{Available open source under \url{https://github.com/uwe-koeckemann/ai4eu-maze-planner/}} illustrates how planning, execution, simulation, and a user
interface can be connected and orchestrated in \ourplatform.  It contains several
loops for task request and achievement, action execution and, state updates.
The topology is illustrated in Figure~\ref{figMazePlanner}. %

The \emph{Graphical User Interface (GUI)} is used to assemble planning problems,
interact with a simulator and request tasks from an executor.  The
\emph{Simulator} simulates action execution and provides state updates to the
GUI and the executor.  A \emph{Planner} receives planning problems and returns
solution plans or failure.  The \emph{Executor} connects these three components
and has several internal loops. If it receives a goal, it will take the last
state provided by the simulator and its operator model to assemble a planning
problem. It then requests a plan from the planner. If no plan is found, failure
is reported to the GUI.  Otherwise, the actions in the plan are placed in a
queue to be sent to the simulator. If the action queue is not empty and
currently no action is running, the executor will send the next action. If an
action is successfully executed (by the simulator) the next action will be
started. If an action fails (e.g., the simulator cannot apply it or it does not
have a model), the rest of the queue is discarded and failure is reported to the
user. If all actions in a queue are successfully executed, success is reported
to the user (via the GUI).

Realizing this solution through \ourplatform\ decouples all components and allows to
replace them by compatible alternatives. In the solution, for instance, the
simulator and executor use the same action model, but can easily be replaced by
more realistic versions. Simulation, e.g., could use a more precise model or
simulate random action failures or external events.
A ROS integration for \ourplatform\ is planned, which enables to exchange the simulator for a real ROS-based robotics environment.
A more sophisticated executor could maintain a time-line representation
to decide when to start actions, how long to wait for them to finish, and to
allow parallel execution via scheduling (see, e.g., \cite[Ch. 4]{ghallab2016automated}).
In this case, a scheduler could be placed between planning and execution in Figure \ref{figMazePlanner}.
Execution could start with an empty action set and learn preconditions and effects from
trial and error.  To realize this, we just need to replace/extend the executor with one
that collects data and can use a learner to extract operators from data.

\subsection{Real time object detection in networked cameras}

Developing a system for Urban Analytics in 10 minutes using the \ourplatform.

This example describes a computer vision application that uses algorithms for object detection in images to develop and deploy a system capable of providing "urban data analytics" in a complex scenario.

We show how to use the popular CNN-based \cite{yolo1} algorithm to survey and monitor a street intersection in a typical urban setting. The enormous potential of \ourplatform\ will be further exploited,  extending the pipeline with one simple component that transforms the scope and aim of the original task, showing that its flexibility and modular design can increase, dramatically, software  productivity.

\begin{figure}[htbp]
    \centering
    \includegraphics[scale=0.52]{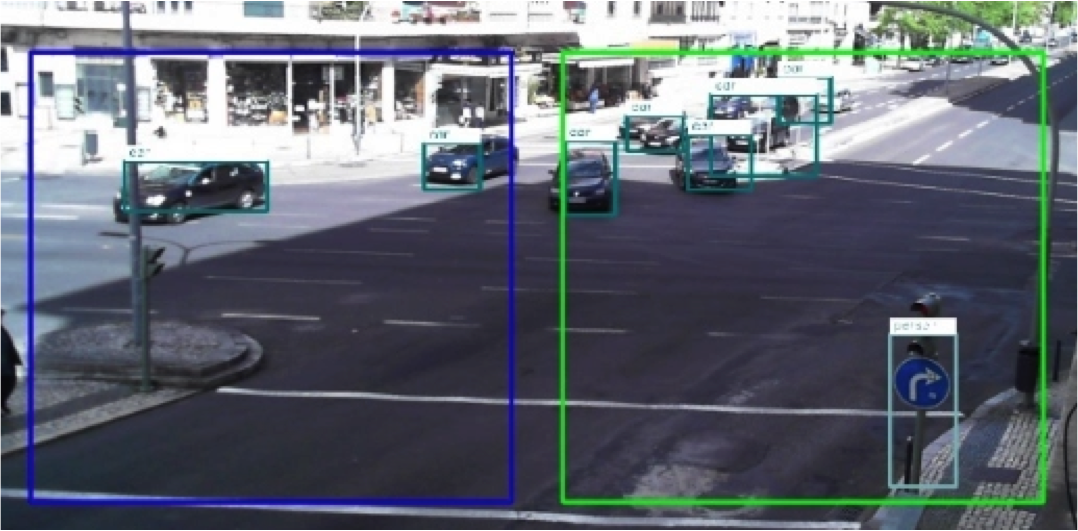}
    \caption{Object detection in urban settings. Each bounding box has a label that  identifies the type of object and the confidence score of the detection. The annotated images are streamed by a webserver.}\label{figyolodetect}
\end{figure}

As Figure \ref{figyolodetect} illustrates, the goal is to acquire images from an IP camera, process it and display the image together with the information of the identified objects. The recognition task is accomplished by the CNN-based detector  YOLOv5\footnote{http://ultralytics.com} that labels each detected object and regresses a bounding box for object location. The annotated image should be accessed through the internet with a browser.

\subsubsection{The \ourplatform\ solution}

Such a system can be easily assembled using the tools available in the \ourplatform, in this case the DesignStudio. Figure \ref{figyolo1} shows the image processing solution where the main components deliver the following tasks:

\begin{figure}[htbp]
    \centering
    \includegraphics[scale=0.62]{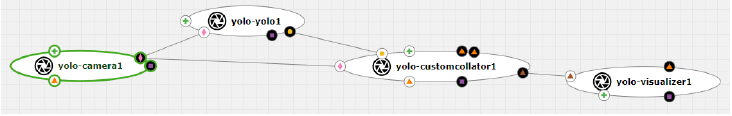}\label{figyolo1}

\begin{tabular}{|r|r l|}
\hline
 \multicolumn{3}{|c|}{Components} \\
 \hline
yolo-camera &  \tt{} rpc& \tt{}Get(Empty) returns (Image);\\
            &\tt{} message &\tt{}Image \{ \\
            &              &\tt{}bytes data = 1;\} \\
       \hline
yolo-yolo &  \tt{}rpc & \tt{}detect(Image) returns (DetectedObjects);\\
&\tt{}message&\tt{}DetectedObjects \{\\
& &\tt{}repeated DetectedObject objects = 1;\}\\
&\tt{}message &\tt{}DetectedObject \{\\
& &   \tt{}string class\_name = 1;\\
& &   \tt{}uint32 class\_idx = 2;\\
& &   \tt{}Point p1 = 3;\\
& &   \tt{}Point p2 = 4;\\
& &   \tt{}double conf = 5;\} \\
&\tt{}message &\tt{}Point \{ \\
& &  \tt{}double x = 1;\\
& &  \tt{}double y = 2;\} \\
       \hline
yolo-visualizer& \tt{} rpc & \tt{}Visualize(ImageWithObjects) returns (Empty);\\
& \tt{}message &\tt{} ImageWithObjects \{ \\
& &\tt{}Image image = 1;\\
& &\tt{}DetectedObjects objects = 2;\}\\
\hline
\end{tabular}
\caption{Pipeline and protobuf definitions for image processing tasks}
\end{figure}

\begin{itemize}
    \item[\textbf{yolo-camera}] Acquires images from an internet camera. The IP and security data (user, pass) are passed as environment variables during deployment and upon request it returns an Image retrieved from the IP camera.
    \item[\textbf{yolo-yolo}]\footnote{https://www.ai4europe.eu/research/ai-catalog/yolo-v5-object-detection} The CNN-based object detector accepts one image as input and outputs a message with the list of detected objects, its location and label  confidence score.
    \item [\textbf{yolo-visualizer}]\footnote {Source for all components:  https://github.com/DuarteMRAlves/yolov5-grpc/} The visualization component deploys a Flask-based web server, and serves a web page that displays the annotated image. The input of this service has two messages: one image and a list of objects.
\end{itemize}

Besides the processing components, this pipeline requires a special node, the ``custom collator", tasked to merging messages coming from different nodes. As described in the table of Figure \ref{figyolo1}, the input to the yolo-visualizer service is a message composed of one image and a list of detected objects  ({\tt ImageWithObjects}), that are originated in different nodes. Thus, the ``custom collator" collects the image from the camera (message {\tt Image}), the output from Yolo (message {\tt DetectedObjects}) and composes a message of type {\tt ImageWithObjects} which will be sent to the Visualizer.

This application is available in \url{http://aiexp.ai4europe.eu}.

\subsubsection{Geo Location and Scaling Up to a Network }

The data collected from street cameras can easily be geo-referenced, anchoring the extracted  "analytics" to global coordinates. Also, by anchoring the detections to global references, we can escalate/fuse this data to a network of similar devices with non overlapping viewpoints. The above system could be deployed seamlessly to any available camera and the the network's output data works as one single "data source".

\begin{figure}[tbph]
    \centering
    \includegraphics[scale=0.42]{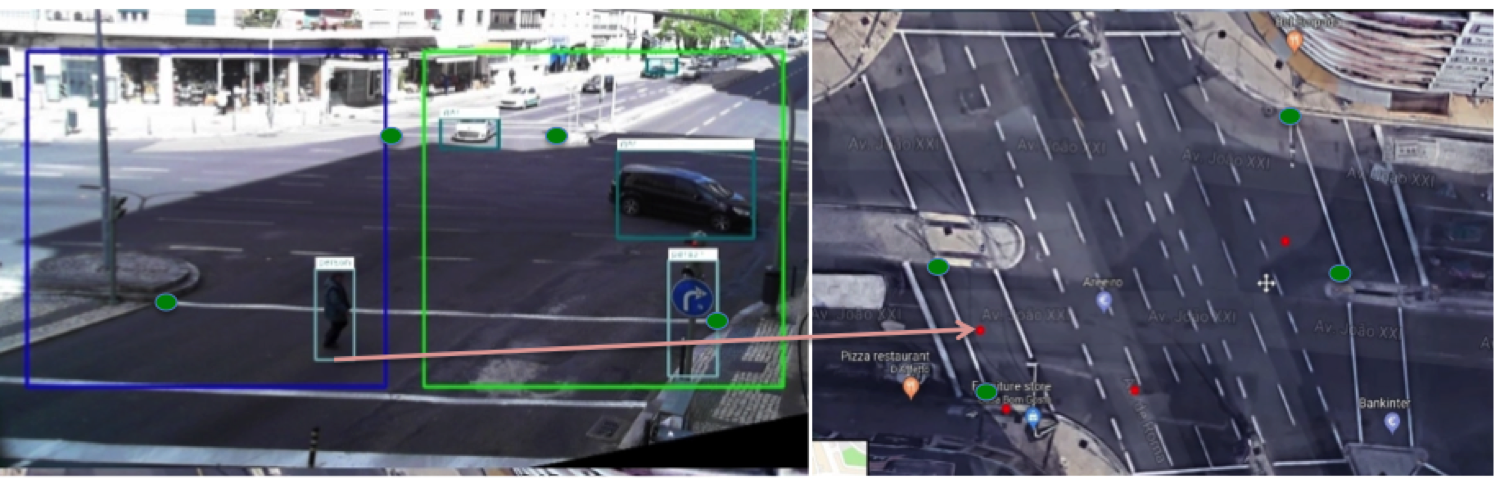}
    \caption{Green dots are used to compute the transformation between cameras. Red dots represent the predicted location of the midpoint of the bounding box low segment}\label{figyologmaps1}
\end{figure}

Assuming the ground is well approximated locally by a plane, there is a projective transformation that maps corresponding points in two images - an \emph{homography}- which can be estimated from a minimum of 4 pairs of non colinear points.

Leveraging services like Google Maps or OpenStreetMaps, we can map the camera image to geo-referenced satellite images of the same area. To estimate the \emph{homography}. In the case illustrated in Figure \ref{figyologmaps1}, the selected pairs of calibration points are identified by the green dots. Of course, the ``flat world" assumption does not hold for vehicles (or pedestrians), particularly if they are close to the camera. However, as we show in the figure (red dots), the midpoint of the bottom line segment of the bounding box is often close to the ground and its mapping is precise enough for the task at hand.

\ifdraft
\else%
In an extension of the above pipeline, we would have a second branch that handles the geo-referenced information.
\fi%
A fully flexible and general structure is easily setup if we introduce a special ``camera" node that simply crawls a website for the adequate satellite image and feeds the ``custom collator" with the corresponding image. Most of the methodologies in computer vision are intuitive to a non-specialist, particularly those involving 3D space, however the maths is often inaccessible to ``lay users/programmers". With this example we show the transformational role that platforms like \ourplatform\ can play, empowering unskilled users with AI technologies that play a key role in their specific domain.

\section{Conclusion and Outlook}
\label{secConclusions}

We described how we create the \ourplatform\
which enables the composition of a broad range of AI applications
based on several extensions of the Acumos framework.
This is the beginning of a long-term effort to create an ecosystem
where modular AI components and visually composed solutions are used
for experimentation, prototyping, and educational purposes
by researchers, industry stakeholders, students, and further interested groups.
In particular the visual composition and a mechanism for finding
matching components for some output port of a component
is intended to lower the barriers for using the system.
Over time, more and more components and useful generic interfaces will be onboarded in the platform
and we foresee that with each addition the system will become more useful for a broader audience.

This work started as a part of the AI4EU H2020 project and will be continued
under the governance of the Eclipse foundation as ``Eclipse Graphene''.%
\footnote{\url{https://projects.eclipse.org/projects/technology.graphene}}
A range of video tutorials is available on YouTube.%
\footnote{\url{https://www.youtube.com/playlist?list=PLL80pOdPsmF6s6P6i2vZNoJ2G0cccwTPa}}

\section*{Acknowledgements}

This work has been supported by the
European Union's Horizon 2020 research and innovation programme under
grant agreement No. 825619 (AI4EU).

\bibliographystyle{plain}
\bibliography{main}

\end{document}